\documentclass[10pt,twocolumn,letterpaper]{article}

\pdfoutput=1

\usepackage{cvpr}
\usepackage{times}
\usepackage{epsfig}
\usepackage{graphicx}
\usepackage{amsmath}
\usepackage{amssymb}

\usepackage{multirow}
\usepackage{bbm}
\usepackage{setspace}
\usepackage{CJKutf8}
\usepackage{nccbbb}

\cvprfinalcopy % *** Uncomment this line for the final submission

\def\cvprPaperID{140} % *** Enter the CVPR Paper ID here

\ifcvprfinal\fi

\newcommand{\figvspace}{\vspace{-0.3cm}}
\newcommand{\figupvspace}{\vspace{-0.5cm}}
\newcommand{\equpvspace}{\vspace{-0.47cm}}
\newcommand{\eqvspace}{\vspace{-0.1cm}}
\newcommand{\tablevspace}{\vspace{-0.6cm}}
\newcommand{\tableupvspace}{\vspace{-0.25cm}}

\begin{document}

%%%%%%%%% TITLE
\title{Beyond triplet loss: a deep quadruplet network for person re-identification}

\author{Weihua Chen$^{1,2}$, Xiaotang Chen$^{1,2}$, Jianguo Zhang$^{3}$, Kaiqi Huang$^{1,2,4}$\\
$^{1}$CRIPAC$ \& $NLPR, CASIA \quad $^{2}$University of Chinese Academy of Sciences\\
$^{3}$Computing, School of Science and Engineering, University of Dundee, United Kingdom\\
$^{4}$CAS Center for Excellence in Brain Science and Intelligence Technology\\
\tt\small{Email:$\{$weihua.chen, xtchen, kqhuang$\}$@nlpr.ia.ac.cn, j.n.zhang@dundee.ac.uk}}

%% omit the following lines up until the closing ``}''.
%% Additional authors and addresses can be added with ``\and'',
%% just like the second author.
%% To save space, use either the email address or home page, not both
%\and

\maketitle
\thispagestyle{empty}

%%%%%%%%% ABSTRACT
\begin{abstract}
Person re-identification (ReID) is an important task in wide area video surveillance which focuses on identifying people across different cameras. Recently, deep learning networks with a triplet loss become a common framework for person ReID. However, the triplet loss pays main attentions on obtaining correct orders on the training set. It still suffers from a weaker generalization capability from the training set to the testing set, thus resulting in inferior performance. In this paper, we design a quadruplet loss, which can lead to the model output with a larger inter-class variation and a smaller intra-class variation compared to the triplet loss. As a result, our model has a better generalization ability and can achieve a higher performance on the testing set. In particular, a quadruplet deep network using a margin-based online hard negative mining is proposed based on the quadruplet loss for the person ReID. In extensive experiments, the proposed network outperforms most of the state-of-the-art algorithms on representative datasets which clearly demonstrates the effectiveness of our proposed method.
\end{abstract}

%%%%%%%%% BODY TEXT
\section{Introduction}

Person re-identification (ReID) is an important task in wide area video surveillance.
The key challenge is the large appearance variations,
usually caused by the significant changes in human body poses, illumination and views.

\begin{figure}[!t]
\centering
\includegraphics[width=1\linewidth]{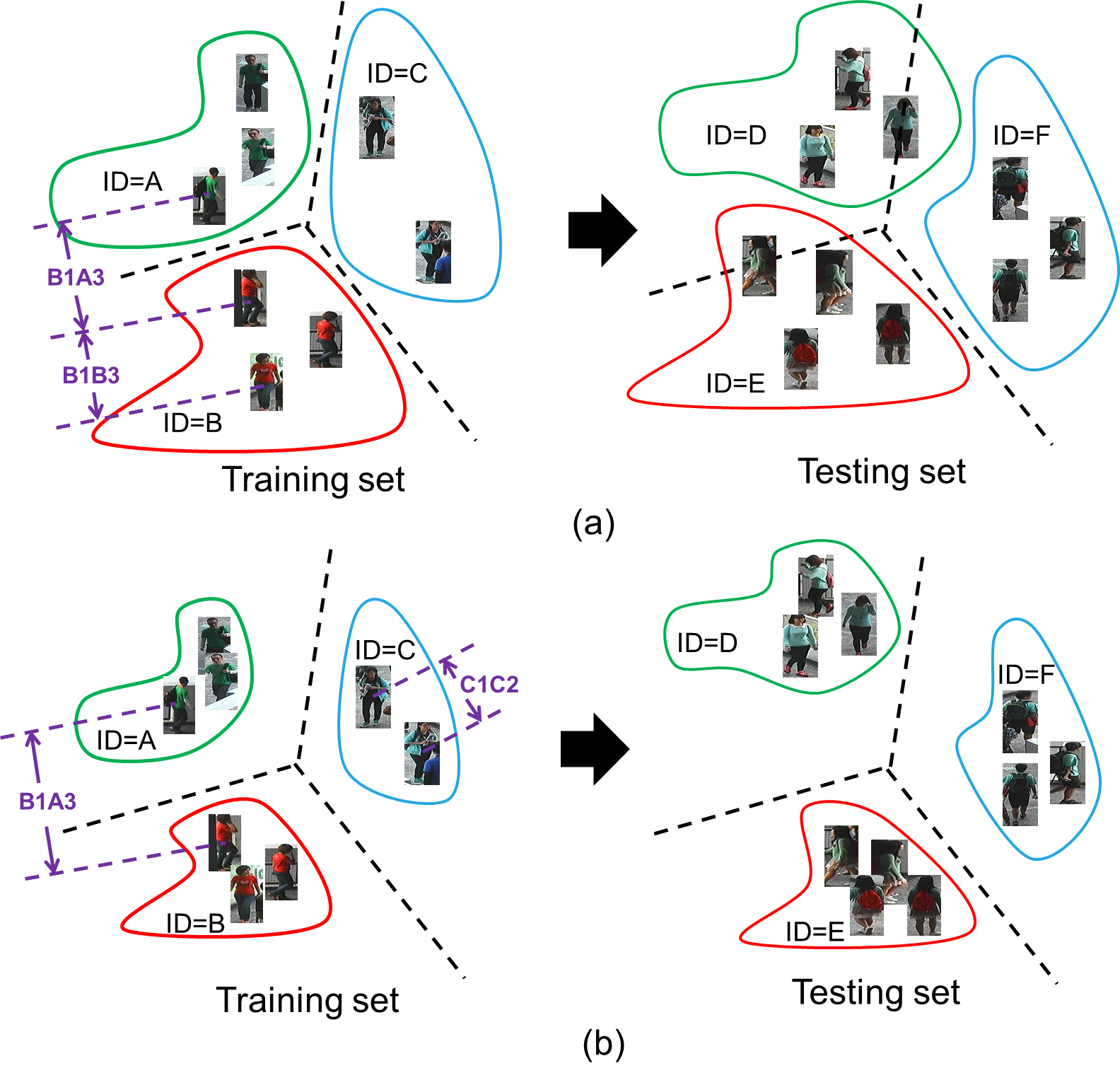}
\figupvspace
\caption{(a) and (b) illustrate the effects of two models (\eg with triplet loss vs. quadruplet loss) learned on the same training set (left) when applied on the same test set (right).
We can see that the model trained in case (b) has output a small intra-class variation and a large inter-class variation, thus tends to perform better on the testing set than the model trained in case (a).}
\label{fig:distribution}
\figvspace
\end{figure}

As person ReID commonly uses the Cumulative Matching Characteristic curve~\cite{hirzer2012avss,karanam2016comprehensive,zheng2016person} for performance evaluation
which follows rank-\emph{n} criteria,
recently deep learning approaches~\cite{ding2015deep,chen2015deep,deepattr2016arXiv,sircir2016cvpr,imptrp2016cvpr}
usually treat the person ReID as a ranking task and apply a triplet loss to address the problem.
The main purpose of the triplet loss is to obtain a correct order for each probe image
and distinguish identities in the projected space.
However, in person ReID the categories (\ie person identities) in the testing set are unseen and have \textit{no overlap} with the training categories.
As shown in Fig.~\ref{fig:distribution} (a), a model learned (\eg typically by a triplet loss) in the training set is specific to the training identities, and performs well in distinguishing these identities. When it is applied on the unseen testing identities, the trained model struggles to be a good performer, showing a weaker generalization capability from training to testing. The underlying reason is that the model trained by a triplet loss would still cause a relatively large intra-class variation\footnote{The category for the intra- and inter- class variations refers to person identities in person ReID.}, which was also observed in~\cite{imptrp2016cvpr}.
It is noted that reducing intra-class variations and enlarging inter-class variations can decrease the generalization error of trained models~\cite{wen2016discriminative}.
We argue that the performance of the triplet loss on the testing set can be improved by further reducing the intra-class variations and enlarging the inter-class variations.
A desired output is shown in Fig.~\ref{fig:distribution} (b).

In this paper, we introduce a \textit{quadruplet} ranking loss, which is modified based on the triplet loss and capable of achieving a smaller intra-class variation and a larger inter-class variation with significant performance on the testing set.
Our designed loss simultaneously considers the following two aspects in one quadruplet: 1) obtaining correct orders for pairs w.r.t the same probe image (\eg $B1B3<B1A3$ in Fig.~\ref{fig:distribution}); and 2) pushing away negative pairs from positive pairs w.r.t different probe images (\eg $C1C2<B1A3$ in Fig.~\ref{fig:distribution}).
The first aspect shares the same idea with the triplet loss and is to keep the correct orders of each probe image in the training set, while the second aspect focuses on further reducing the intra-class variations and enlarging the inter-class variations.
%With the help of the second aspect, our loss achieves a requirement that all positive pairs should have smaller distances than negative pairs.
%In other words, the global minimum inter-class distance has to be larger than the global maximum intra-class distance in our loss.
The balance of these two aspects is controlled implicitly by two margins.
It is worth mentioning that, the second aspect is not necessary for a good result on the training set, but we argue it's helpful to enhance the generalization ability of the trained models on the testing set.
Experiments in Section~\ref{sec:experiments} demonstrate that this design can produce larger inter-class variations and smaller intra-class variations, and thus lead to a better performance on the testing set.

In addition to a triplet loss, some deep learning methods~\cite{Li/cvpr2014deepreid,ahmed/cvpr2015improved,yi/icpr2014deep,personnet2016arXiv,deepattr2016arXiv,sircir2016cvpr}
address the person ReID problem from the classification aspect and adopt a binary classification loss to train their models.
To justify the proposed loss, we present a theoretical analysis of the relationships of three different losses: our quadruplet loss, the triplet loss and the commonly used binary classification loss. To the best of our knowledge, this is the first detailed study of such relationships in a unified view for person ReID.

Meanwhile, we propose a quadruplet deep network based on our quadruplet loss.
In the proposed network, the input sample is a quadruplet.
In practice, even for a small dataset, it can produce an overwhelming number of quadruplet samples.
Selecting suitable samples for training a deep net is a big challenge.
We introduce a margin-based online hard negative mining to select hard samples to train the model.
Our algorithm adaptively sets the margin threshold according to the trained model,
and uses this margin threshold to automatically select hard samples.

In summary, our contributions are four-fold:
1) a quadruplet loss, with \textit{strong} and \textit{weak} push strategies;
2) a quadruplet deep network with a margin-based online hard negative mining strategy;
3) a theoretical and insightful analysis of loss relationships, putting different losses in a unified view;
4) significant performance on representative datasets (\eg CUHK03, CUHK01 and VIPeR), being superior to most of the state-of-the-art methods.

%The rest of the paper is organized as follows. Related work is introduced in Section \ref{sec:relatedwork}.
%The quadruplet loss and the proposed deep network are described in Section \ref{sec:quadrupletloss}.
%Section \ref{sec:relationship} presents the relationship between our loss and the binary classification loss.
%The experimental results and the conclusion are provided in Section \ref{sec:experiments} and Section \ref{sec:conclusion}.

%\begin{figure}[!t]
%\centering
%\includegraphics[width=1\linewidth]{./fig/distribution3.png}
%\caption{(a) and (b) illustrate the two models trained by the triplet loss and our quadruplet loss respectively.}
%\label{fig:distribution3}
%\vspace{-1.1em}
%\end{figure}

\section{Related work}
\label{sec:relatedwork}

Most of existing methods in person ReID focus on either feature extraction
\cite{Zhao/cvpr2014midlevel,su2015multi,hierarchical2016cvpr,gou2016person,yang2017aaai}, or similarity measurement
\cite{Li/cvpr2013locally,Shen/iccv2015structure,Liao/iccv2015psd,yang2016metric}.
Person image descriptors commonly used include color histogram \cite{Koestinger/cvpr2012scale,Li/cvpr2013locally,xiong/eccv2014person},
local binary patterns \cite{Koestinger/cvpr2012scale},
Gabor features \cite{Li/cvpr2013locally}, and etc.,
which show certain robustness to the variations of poses, illumination and viewpoints.
For similarity measurement, many metric learning approaches are proposed to learn a suitable metric,
such as locally adaptive decision functions \cite{li/cvpr2013learning},
local fisher discriminant analysis \cite{pedagadi/cvpr2013local},
cross-view quadratic discriminant analysis \cite{liao2015person},
and etc.
%A few of them \cite{xiong/eccv2014person,paisitkriangkrai/cvpr2015learning} learn a combination of multiple metrics.
However, manually crafting features and metrics are usually not optimal to cope with large intra-class variations.

Since feature extraction and similarity measurement are independent,
the performance of the whole system is often suboptimal
compared with an end-to-end system using CNN that can be globally optimized via back-propagation.
With the development of deep learning and the increasing availability of datasets,
the handcrafted features and metrics struggle to keep top performance widely, especially on large scale datasets.
Alternatively, deep learning
is attempted for person ReID to automatically learn features and metrics
\cite{Li/cvpr2014deepreid,ahmed/cvpr2015improved,sircir2016cvpr,chen2017aaai,dangwei2017cvpr}.
Some deep methods~\cite{ding2015deep,chen2015deep,imptrp2016cvpr} consider person ReID as a ranking issue.
For example, Ding \etal~\cite{ding2015deep} use a triplet loss to get the relative distance between images.
Chen \etal~\cite{chen2015deep} design a ranking loss which minimizes the cost corresponding to the sum
of the gallery ranking disorders.
Our method also solves person ReID on ranking aspect and introduces a quadruplet loss
which enlarges inter-class variations and reduces intra-class variations.

Meanwhile, there are approaches
~\cite{Li/cvpr2014deepreid,ahmed/cvpr2015improved,yi/icpr2014deep,personnet2016arXiv,sircir2016cvpr,gated2016cvpr,lstm2016cvpr}
which tackle the person ReID problem from the classification aspect.
Some of them adopt a softmax layer with the cross-entropy loss in their networks~\cite{Li/cvpr2014deepreid,ahmed/cvpr2015improved,personnet2016arXiv}.
The cross-entropy loss can well represent the probability that the two images in the pair are of the same person or not.
Others~\cite{sircir2016cvpr,gated2016cvpr,lstm2016cvpr} import a margin-based loss (\eg a contrastive loss~\cite{ctsloss2006cvpr}),
which builds a margin to keep the largest separation between positive and negative pairs.
For instance, Varior~\etal~\cite{lstm2016cvpr} design a siamese LSTM architecture with a contrastive loss.
In Section \ref{sec:relationship}, we analyse the relationship between different losses
which justifies the proposal of our quadruplet loss.

It is worth mentioning that there are two deep methods (DeepLDA~\cite{LDA2016arXiv} and ImpTrpLoss~\cite{imptrp2016cvpr})
which also manage to reduce the intra-class variations like us in person ReID.
DeepLDA~\cite{LDA2016arXiv} imports a LDA objective function using fisher vectors.
However, it pays all its attentions on the intra- and inter- class variations and partly ignore the relative relationships between pairs.
Our quadruplet loss is expanded from the triplet loss which reserves the relative relationships in the trained model.
ImpTrpLoss~\cite{imptrp2016cvpr} imports an additional constraint in the traditional triplet loss,
which limits the distances of positive pairs to be smaller than a pre-defined value,
while in our method, the new constraint comes from the pairs with different probe images.

What's more, there are works exploring effective sampling schemes~\cite{ahmed/cvpr2015improved,ding2015deep} in person ReID.
Ahmed~\etal~\cite{ahmed/cvpr2015improved} iteratively fine-tune their models with hard negative samples selected by a previous trained model, which is an offline hard negative mining method.
While Ding~\etal~\cite{ding2015deep} design a predefined triplet generation scheme. In each iteration, they randomly select a small number of classes (persons) and generate triplets using only those images.
However, these methods can't select samples adaptively according to the trained model.
The margin threshold in our method is adaptively set according to the trained model,
which can be used to automatically select hard samples.

\begin{figure}[!t]
\centering
\includegraphics[width=1\linewidth]{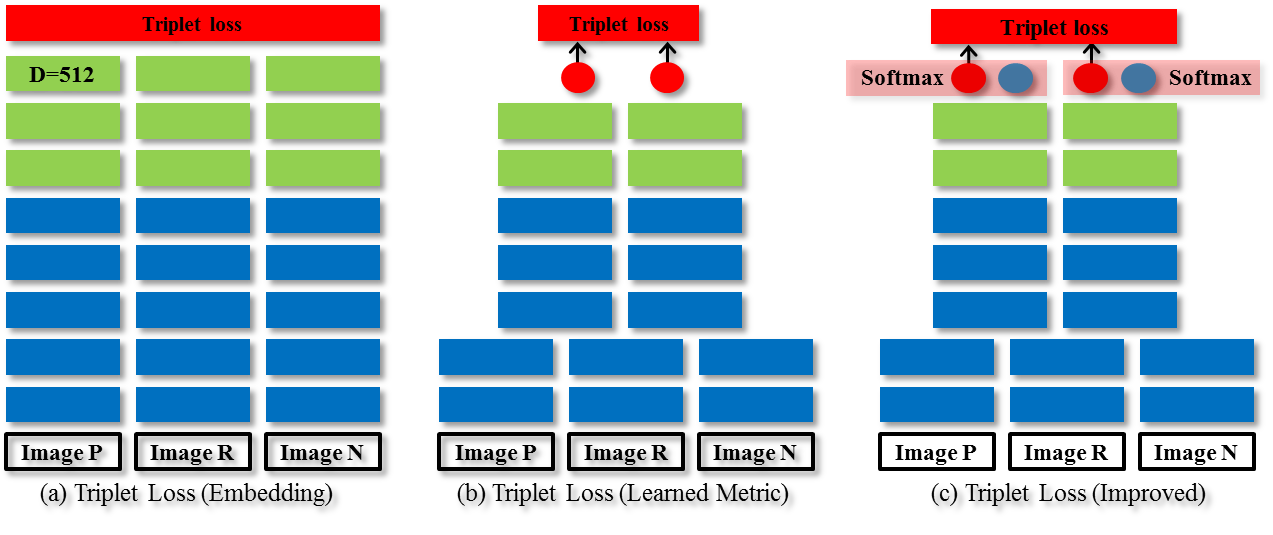}
\figupvspace
\caption{Three networks with different losses. (a) the triplet loss with Euclidean distance~\cite{schroff2015facenet}; (b) the triplet loss with learned metric~\cite{sircir2016cvpr}; (c) our improved net with the normalised triplet loss.}
\label{fig:nets}
\figvspace
\end{figure}

\section{The proposed approach}
\label{sec:quadrupletloss}

Our quadruplet is designed based on the commonly used triplet loss.
So, in this section, we first introduce the triplet loss and then present our quadruplet loss.
The proposed network with the margin-based online hard negative mining would be introduced at last.

\subsection{The triplet loss}
\label{ssec:tripletloss}

The triplet loss~\cite{schroff2015facenet} is normally trained on a series of triplets $\{x_i,x_j,x_k\}$,
where $x_i$ and $x_j$ are images from the same person, and $x_k$ is from a different person.
The triplet loss is designed to keep $x_i$ closer to $x_j$ than $x_k$,
and widely used in many areas, such as image retrieval~\cite{Wang2014cvpr}, face recognition~\cite{schroff2015facenet} and person re-identification~\cite{ding2015deep,imptrp2016cvpr}.
It is formulated as following:

\equpvspace
\begin{equation}
L_{trp}\!=\!\sum_{i\!,j\!,k}^N{[\|\!f(\!x_i\!)\!-\!f(\!x_j\!)\|_2^2\!-\!\|\!f(\!x_i\!)\!-\!f(\!x_k\!)\!\|_2^2\!+\!\alpha_{trp}]_{+}}
\label{eq:embedtripletloss}
\eqvspace
\end{equation}
where $[z]_{+}=max(z,0)$, and $f(x_i),f(x_j),f(x_k)$ mean features of three input images.

In most cases, the image feature $f$ is well normalized during training.
The threshold $\alpha_{trp}$ is a margin that is enforced between positive and negative pairs.
The related network is shown in Fig.~\ref{fig:nets} (a).
In Eq.~\ref{eq:embedtripletloss}, the triplet loss adopts the Euclidean distance to measure the similarity of extracted features from two images.
We replace the Euclidean distance with a learned metric $g(x_i,x_j)$, similar to Wang \etal~\cite{sircir2016cvpr},
which can effectively model the complex relationships between the gallery and probe images,
and can be more robust to appearance changes across cameras.
The loss with the learned metric is formulated as:

\equpvspace
\begin{equation}
L_{trp}=\sum_{i,j,k}^N{[g(x_i,x_j)^2-g(x_i,x_k)^2+\alpha_{trp}]_{+}}
\label{eq:learnedtripletloss}
\eqvspace
\end{equation}

In Eq.~\ref{eq:embedtripletloss}, $f(x_i)$ is well normalized and keeps $\|f(x_i)-f(x_j)\|_2$ ranging in [0,1].
But in Eq.~\ref{eq:learnedtripletloss}, $g(x_i,x_j)$ is a value instead of a vector.
Wang \etal~\cite{sircir2016cvpr} use a fully connected layer with a one-dimensional output to learn the value $g(x_i,x_j)$ as shown Fig.~\ref{fig:nets} (b).
It would cause the value $g(x_i,x_j)$ can't maintain the range of [0,1],
and partly invalidate the margin threshold $\alpha_{trp}$.
For example, no matter how large the threshold $\alpha_{trp}$,
the model can simultaneously multiply $g(x_i,x_j)$ and $g(x_i,x_k)$ by an appropriate value to meet the requirement of the margin threshold.

So in Section~\ref{ssec:quadrupletloss}, we first introduce our improvement on the triplet loss and then present our quadruplet loss.

\subsection{The quadruplet loss}
\label{ssec:quadrupletloss}

At beginning, we first propose an improvement to handle the lack of normalization in Fig.~\ref{fig:nets} (b).
A fully connected layer with a two-dimensional output is adopted in our net as shown in Fig.~\ref{fig:nets} (c).
As $g(x_i,x_j)$ represents the distance of two images, the larger $g(x_i,x_j)$ is, the more dissimilar two images are.
The value $g(x_i,x_j)$ should be positively correlated with the probability of dissimilarity of two images.
Thus we assumed that one of two dimensions in our fully connected layer can represent probabilities of dissimilarity of two images to some extent.
A softmax layer is adopted to normalize the two dimensions.
Then one dimension, \ie the one representing the dissimilarity of two images (red point in Fig.~\ref{fig:nets} (c)),
is used to serve as $g(x_i,x_j)$ to be sent to the loss and be trained.
As a result, the value $g(x_i,x_j)$ can be well normalized and range in [0,1] which ensures the effectiveness of the margin threshold $\alpha_{trp}$.

Additionally we have explored to import a softmax loss in Fig.~\ref{fig:nets} (c) after the final fully connected layer, which can enhance the two outputs on representing probabilities of similarity and dissimilarity of two images to some extent.
Its influence is discussed in Section~\ref{ssec:experiment1}
with the comparison between \textit{Triplet(Improved w/o sfx)} and \textit{Triplet(Improved)}.

\begin{figure*}
\centering
\includegraphics[width=0.9\linewidth]{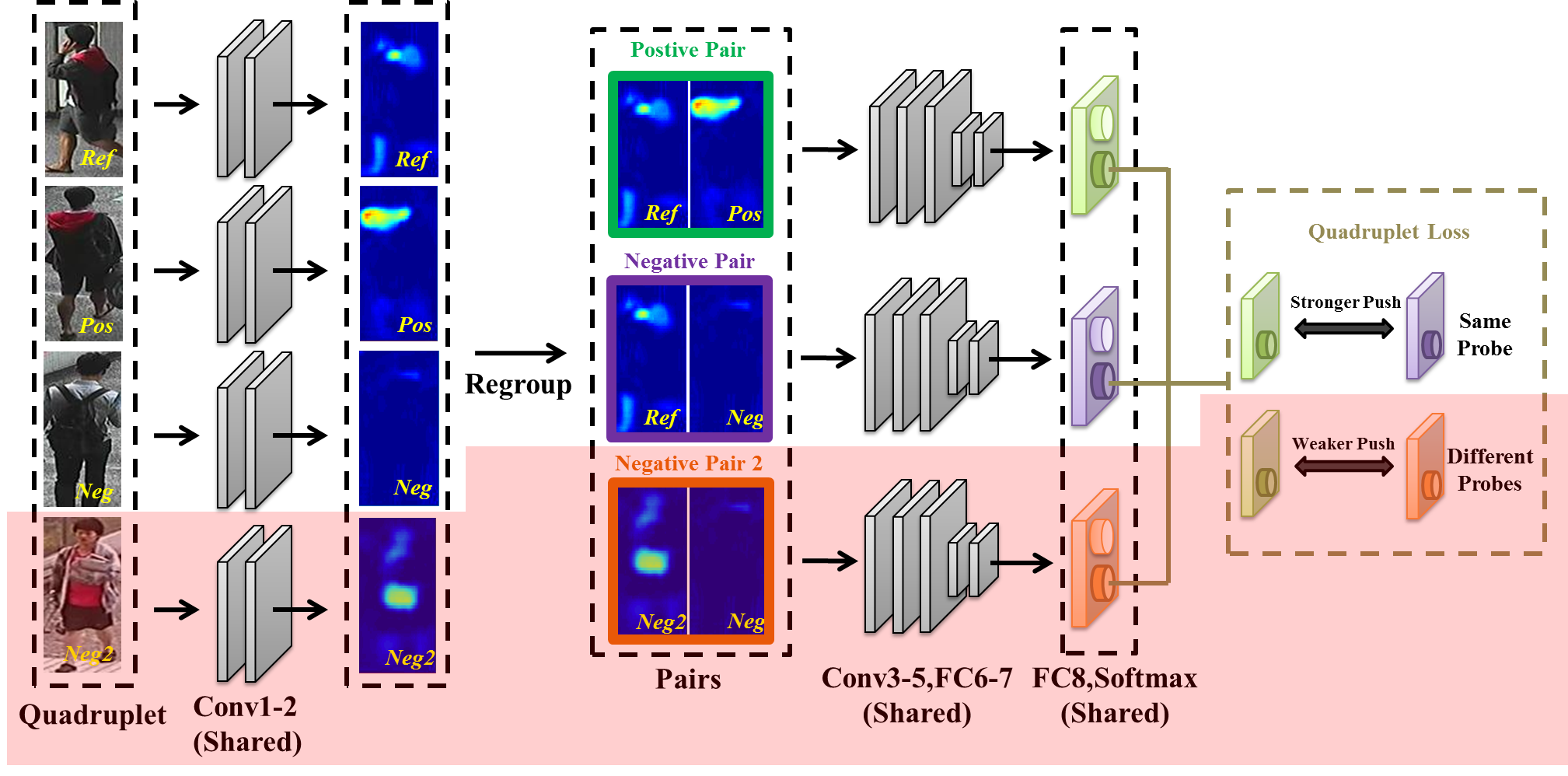}
\vspace{-0.1cm}
\caption{The framework of the proposed quadruplet deep network. The red shadow region indicates elements of the new constraint.}
\label{fig:framework}
\figvspace
\end{figure*}

From Eq.~\ref{eq:learnedtripletloss}, it's evident that the triplet loss trains the model only based on the relative distances
between positive and negative pairs w.r.t the same probe images.
Our quadruplet loss introduces a new constraint
which pushes away negative pairs from positive pairs w.r.t different probe images.
The quadruplet loss is as below:

\equpvspace
\begin{equation}
\begin{aligned}
& L_{quad}=\sum_{i,j,k}^N{[g(x_i,x_j)^2-g(x_i,x_k)^2+\alpha_1]_{+}} \\
& \phantom{L_{quad}=}+\sum_{i,j,k,l}^N{[g(x_i,x_j)^2-g(x_l,x_k)^2+\alpha_2]_{+}} \\
& \phantom{XXXXX}s_i=s_j,s_l\neq s_k,s_i\neq s_l,s_i\neq s_k
\label{eq:quadrupletloss}
\end{aligned}
\eqvspace
\end{equation}
where $\alpha_1$ and $\alpha_2$ are the values of margins in two terms and $s_i$ refers to the person ID of image $x_i$.

The first term is the same as Eq.~\ref{eq:learnedtripletloss}.
It focuses on the relative distances between positive and negative pairs w.r.t the same probe images.
The second term is the new constraint which considers the orders of positive and negative pairs with different probe images.
With the help of this constraint, the minimum inter-class distance is required to be larger than the maximum intra-class distance regardless of whether pairs contain the same probe.

As mentioned above, the first term aims to obtain the correct orders with the same probe in training data.
%But even its optimal solution can't guarantee the correctness of orders in testing data.
The second term provides a help from the perspective of orders with different probe images.
It can further enlarge the inter-class variations and improve the performance on the testing data.
Though it's a useful auxiliary term, it should not lead the training phase and be considered as equally important as the first term.
Therefore, we treat the two terms differently in Eq.~\ref{eq:quadrupletloss}.
We adopts the margin thresholds to determine the balance of two terms in our loss instead of using weights.
We require that the margin between pairs with the same probe should be large enough to maintain the main constraint.
And the second term could hold a smaller margin to achieve a relatively weaker auxiliary constraint.
So in our method, $\alpha_1$ is set to be larger than $\alpha_2$.

The framework of our network using the quadruplet loss is shown in Fig.~\ref{fig:framework}.
%Our network takes the four images, each of which is processed through two convolutional layers to generate feature maps. These feature maps are then regrouped into three pairs including one positive pair and two negative pairs. For each pair, the two feature maps are stacked on the channel dimension and formed into paired features. The three types of paired features are fed to three convolutional layers and three fully connected layers. The FC8 is the fully connected layer with a two-dimensional output. The input of our quadruplet loss is the three 2-dimensional outputs, corresponding to the three types of paired features.
The architecture without the red shadow region is the network in Fig.~\ref{fig:nets} (c).
After bringing in the new constraint, the architecture changes from a triplet network to a quadruplet network.
The quadruplet network not only treats positive and negative pairs differently as the triplet loss does,
but also distinguishes two pairs on whether the probe images are same or not.
For the pairs from the same probe, the quadruplet loss produces a strong push between positive and negative pairs,
while for those with different probes, our loss provides a relatively weaker push to reduce the inter-class variations.

\subsection{Margin-based online hard negative mining}
\label{ssec:onlinehnm}

%Assume a training ReID dataset contains $N$ people,
%and each person has one probe image and one gallery image.
%When using a triplet network, the total number ($N*(N-1)$) of triplet samples is overwhelming
%including many easy samples and a small number of hard samples.
%However, in our quadruplet network, the number of samples is much larger than that in the triplet network
%\footnote{The samples are quadruplets and the total number is $N*(N-1)*(N-1)$}.
%Hence, a good sampling scheme is much more important in ours than in the triplet network.

As we know, the margin thresholds are to confine the distance between positive and negative pairs in a quadruplet sample.
In Eq.~\ref{eq:embedtripletloss}, Schroff~\etal~\cite{schroff2015facenet}
select the samples which hold a smaller distance than the margin threshold as hard samples to achieve an online hard negative mining.
However, it's hard to predefine a suitable margin threshold.
A small threshold would result in few hard samples.
As only hard samples are feedback to train the model,
few hard samples would cause a slow convergence and easily lead the model to a suboptimal solution.
In contrast, a large threshold would produce too much hard training samples to cause over fitting.
Our algorithm manages to adaptively set the margin threshold according to the trained model,
and use this margin threshold to select hard samples.

The main idea behind our adaptive margin is the avoidance of the over- or under-sampling problems mentioned above
with the assumption that the distances (or similarities) between features of the same class (\ie positive pairs)
or different classes (\ie negative pairs) are samples from two distinct distributions,
\ie the positive pair distance distribution and the negative pair distance distribution.
The adaptive margin threshold is substantially used to express the average distance of the two distributions,
which should have a positive relationship with the average distance.
So we use the average distance of two distinct distributions to adaptively represent our margin thresholds.

\equpvspace
\begin{equation}
\begin{aligned}
& \alpha = w(\mu_n-\mu_p) \\
& \phantom{\alpha} = w(\frac{1}{N_n}\!\sum_{i\!,k}^N{g(\!x_i\!,\!x_k\!)\!^2}\!-\!\frac{1}{N_p}\!\sum_{i\!,j}^N{g(\!x_i\!,\!x_j\!)\!^2}) \\
& \phantom{XXXXX}s_i=s_j,s_i\neq s_k\\
\label{eq:marginthrd}
\end{aligned}
\vspace{-0.7cm}
\end{equation}
where $\mu_p$ and $\mu_n$ are mean values of two distributions.
%The threshold $\theta$ is to control the valid range of the average distance and set to 0.5 in our experiments.
$N_p$ and $N_n$ are numbers of positive and negative pairs respectively,
and $w$ is the correlation coefficient.
We set $w=1$ for $\alpha_1$ and $w=0.5$ for $\alpha_2$ in Eq.~\ref{eq:quadrupletloss}.
%From on Eq.~\ref{eq:marginthrd}, the margin threshold is self-adaptive based on the two distributions of the trained model.

%As a result, our pairrank loss can be rewritten as:
%
%\begin{equation}
%\begin{aligned}
%& L_{pk}=\sum_{i,j,k}^N{[g(x_i,x_j)^2-g(x_i,x_k)^2+} \\
%& \phantom{XXXXX}(\frac{1}{N_n}\sum_{i,k}^N{g(x_i,x_k)}-\frac{1}{N_p}\sum_{i,j}^N{g(x_i,x_j)})]_{+} \\
%& \phantom{L_{pk}=}+\sum_{i,j,k,l}^N{[g(x_i,x_j)^2-g(x_l,x_k)^2} \\
%& \phantom{XXXXX}+\frac{1}{2}(\frac{1}{N_n}\sum_{i,k}^N{g(x_i,x_k)}-\frac{1}{N_p}\sum_{i,j}^N{g(x_i,x_j)})]_{+} \\
%& \phantom{XXXXX}s_i=s_j,s_l\neq s_k,s_i\neq s_l,s_i\neq s_k
%\label{eq:addconstraint}
%\end{aligned}
%\end{equation}

In the implementation, computing the mean of two distributions for each iteration is time consuming.
We use the mean of two distributions in each batch instead.
Assuming the batch size is $M$, and $N_p$ and $N_n$ would be set to $M$ and $2M$ respectively.
Given that we use stochastic gradient descent (SGD) for the optimisation process,
we need to derive the gradient of our loss function, as follows:

\equpvspace
\begin{equation}
\begin{aligned}
& \frac{\partial L_{quad}}{\partial g(x\!_i\!,\!x\!_j)}\!=\!(\!2\!-\!\frac{2}{M}\!)g(\!x\!_i\!,\!x\!_j\!)\bbbone[g(\!x\!_i\!,\!x\!_k\!)\!^2\!-\!g(\!x\!_i\!,\!x\!_j\!)\!^2\!<\!\max\!(\mu\!,\!0)]\\
& \phantom{XX}+\!(\!2\!-\!\frac{1}{M}\!)g(\!x\!_i\!,\!x\!_j\!)\bbbone[g(\!x\!_l\!,\!x\!_k\!)\!^2\!-\!g(\!x\!_i\!,\!x\!_j\!)\!^2\!<\!\frac{\!\max\!(\mu\!,\!0)\!}{\!2\!}]\\
& \frac{\partial L_{quad}}{\partial g(x\!_i\!,\!x\!_k)}\!=\!(\!-\!2\!+\!\frac{\!3\!}{\!2M\!}\!)g(\!x\!_i\!,\!x\!_k\!)\bbbone[g(\!x\!_i\!,\!x\!_k\!)\!^2\!-\!g(\!x\!_i\!,\!x\!_j\!)\!^2\!<\!\max\!(\mu\!,\!0)]\\
& \frac{\partial L_{quad}}{\partial g(x\!_l\!,\!x\!_k)}\!=\! (\!-\!2\!+\!\frac{\!3\!}{\!2M\!}\!)g(\!x\!_l\!,\!x\!_k\!)\bbbone[g(\!x\!_l\!,\!x\!_k\!)\!^2\!-\!g(\!x\!_i\!,\!x\!_j\!)\!^2\!<\!\frac{\!\max\!(\mu\!,\!0)\!}{\!2\!}]\\
& \phantom{XXXXX}s_i=s_j,s_l\neq s_k,s_i\neq s_l,s_i\neq s_k
\label{eq:gradloss}
\end{aligned}
\eqvspace
\end{equation}
where $\mu=\mu_n-\mu_p$, and $\bbbone[a]$ is an indicator function with value 1 when $a$ is true, otherwise 0.

Thus the margin threshold is self-adaptive based on the two distributions of the trained model.
During each iteration, only the samples holding smaller distances than the average are selected and back propagated,
which are considered as hard samples in current trained model.
%As only a nearly half number of samples are selected for updating the model, the backward pass is no more expensive than before.

%Besides the online hard negative mining in our quadruplet loss, for the softmax layer in Fig.~\ref{fig:framework}, we compute the loss on all samples in batch, rank and select the top-$k$ as hard samples to perform the feedback, which is similar to Shrivastava~\etal~\cite{shrivastava2016training}. Thus, our network is a purely online form and select hard samples based on the adaptive margin.

\section{Relationships of different losses}
\label{sec:relationship}

In this section, we theoretically discuss the relationships of our quadruplet loss, the triplet loss and the traditional binary classification
loss.

Under a learned metric, the triplet loss and our quadruplet loss can be formulated as Eq.~\ref{eq:learnedtripletloss} and Eq.~\ref{eq:quadrupletloss} respectively.
For the binary classification aspect, either the cross-entropy loss~\cite{Li/cvpr2014deepreid,ahmed/cvpr2015improved}
or the contrastive loss~\cite{gated2016cvpr,lstm2016cvpr} can be used as a binary classification loss.
The cross-entropy loss can well represent the probability that the two images in the pair are of the same person or not.
However, it can't obtain a largest separation between positive and negative pairs due to lack of the margin threshold.
The margin in the contrastive loss can partly enhance the generalization ability of the classifier from the training set to the testing set.
Because in general the larger the margin, the lower the generalization error of the classifier~\cite{cortes1995support}.
So in this section, we mainly compare our quadruplet loss with the contrastive loss, which contains a margin threshold consistently with ours.
The contrastive loss can be formulated as follows:

\equpvspace
\begin{equation}
\begin{aligned}
& L_{cts}=\sum_{i,j}^N{[y_{ij}d+(1-y_{ij})max(0,\alpha_{cts}-d)]} \\
& \phantom{XXXXXXX}d=\|f(x_i)-f(x_j)\|^2_2
\label{eq:contrastloss}
\end{aligned}
\eqvspace
\end{equation}
where $y=1$ for positive pairs, and $y=0$ for negative ones, and $\alpha_{cts}$ is the margin threshold.
When a learned metric $g(x_i,x_j)$ is applied, the loss becomes:

\equpvspace
\begin{equation}
\begin{aligned}
L_{\!cts\!}\!=\!\sum_{i\!,j}^N\!{[y_{i\!j}g(\!x\!_i\!,\!x\!_j\!)^2\!+\!(\!1\!-\!y_{i\!j}\!)max(0,\alpha_{\!cts\!}\!-\!g(\!x\!_i\!,\!x\!_j\!)^2)]}
\label{eq:learnedcontrastloss}
\end{aligned}
\eqvspace
\end{equation}

\begin{figure}[!t]
\centering
\includegraphics[width=1\linewidth]{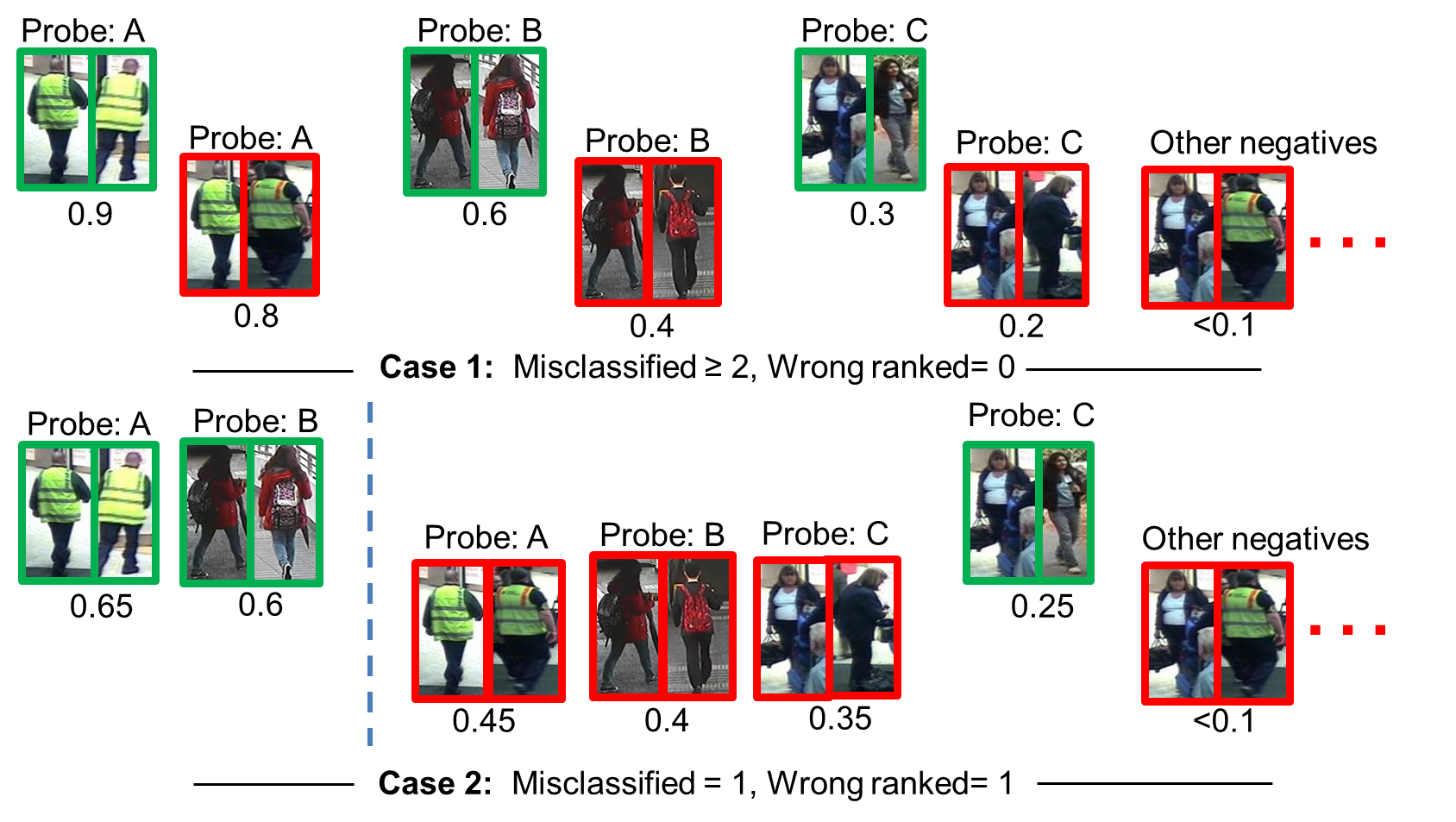}
\figupvspace
\caption{The binary classification loss prefers to
training a lower misclassification rate model like Case 2 rather than Case 1,
which imports a wrong order (Probe C).
And it is an undesired locally optimal solution for person ReID. (Best viewed in color)}
\label{fig:distribution2}
\figvspace
\end{figure}

The input of the contrastive loss in Eq.~\ref{eq:learnedcontrastloss} is doublets $\{(x_i,x_j),y\}$,
while the training samples sent to our quadruplet loss in Eq.~\ref{eq:quadrupletloss} are quadruplets $\{x_i,x_j,x_k,x_l\}$.
If we want to compare Eq.~\ref{eq:learnedcontrastloss} and Eq.~\ref{eq:quadrupletloss}, we have to keep their inputs consist.
Therefore, we manage to transform the doublet samples into quadruplets.
In Eq.~\ref{eq:learnedcontrastloss}, the input sample is a doublet (\ie a pair) containing two images.
We assume that the batch has $M$ samples, which contains ($N_p\!=\!a$) positive pairs and ($N_n\!=\!M\!-\!a$) negative doublets.
During the transformation, we change the input sample into two doublets including a positive doublet and a negative doublet.
As a result, the transformed sample contains four images from two doublets.
Both positive and negative doublets are selected from $M$ original samples,
so that the images used in the training of this batch are not changed and no additional image is imported during the transformation.
The only change is the increased frequency of usages of image pairs.
We assume that this increased frequency would make no difference on performance as long as the usage ratio of positive and negative pairs are unchanged\footnote{In this batch, we can set $a=M/2$ to keep the ratio consistent.}.
The batch size would become $a(M-a)$,
and we adopt ``$\Rightarrow$'' to express the transformation.
For a batch, the contrastive loss can be formulated as following:

\equpvspace
\begin{equation}
\begin{aligned}
& L_{\!cts\!}\!=\!\sum_{(\!i\!,j\!)}^M\!{[y_{i\!j}g(\!x\!_i\!,\!x\!_j\!)^2\!
+\!(\!1\!-\!y_{i\!j}\!)max(0,\alpha_{\!cts\!}\!-\!g(\!x\!_i\!,\!x\!_j\!)^2)]}
\\
& \phantom{L_{\!cts\!}}\!=\!\sum_{(\!i\!,j\!)|{s_i\!=\!s_j}}^a\!{[g(\!x\!_i\!,\!x\!_j\!)^2\!]}
+\!\sum_{(\!i\!,j\!)|{s_i\!\neq\!s_j}}^{M-a}\!{[max(0,\alpha_{\!cts\!}\!-\!g(\!x\!_i\!,\!x\!_j\!)^2)]}
\\
& \phantom{L_{\!cts\!}}\!\Rightarrow\!\sum_{(\!i\!,j\!,l\!,k\!)|{s_i\!=\!s_j\!,s_l\!\neq\!s_k}}^{a(M-a)}\!{[g(\!x\!_i\!,\!x\!_j\!)^2\!
+\!max(0,\alpha_{\!cts\!}\!-\!g(\!x\!_l\!,\!x\!_k\!)^2)]}
\\
& \phantom{L_{\!cts\!}}\!=\!\sum_{(\!i\!,j\!,l\!,k\!)|{s_i\!=\!s_j\!,s_l\!\neq\!s_k}}^{a(M-a)}\!
{max[g(\!x\!_i\!,\!x\!_j\!)^2\!,\!g(\!x\!_i\!,\!x\!_j\!)^2\!+\!\alpha_{\!cts\!}\!-\!g(\!x\!_l\!,\!x\!_k\!)^2]}
\\
& \phantom{L_{\!cts\!}}\!=\!\sum_{(\!i\!,j\!,l\!,k\!)|{s_i\!=\!s_j\!=\!s_l\!,s_l\!\neq\!s_k}}^{b}\!
{max[g(\!x\!_i\!,\!x\!_j\!)\!^2\!,\!g(\!x\!_i\!,\!x\!_j\!)\!^2\!+\!\alpha_{\!cts\!}\!-\!g(\!x\!_l\!,\!x\!_k\!)\!^2]}
\\
& \phantom{L_{\!cts\!}}\!+\!\sum_{(\!i\!,j\!,l\!,k\!)|{s_i\!=\!s_j\!,s_i\!\neq\!s_l\!\neq\!s_k}}^{a(M-a)-b}\!
{max[g(\!x\!_i\!,\!x\!_j\!)\!^2\!,\!g(\!x\!_i\!,\!x\!_j\!)\!^2\!+\!\alpha_{\!cts\!}\!-\!g(\!x\!_l\!,\!x\!_k\!)\!^2]}
\label{eq:convertsets}
\end{aligned}
\eqvspace
\end{equation}
where $s_i$ indicates the person ID of image $x_i$.

In Eq.~\ref{eq:convertsets}, the contrastive loss is split into two terms after the transformation.
The first term focuses on two pairs which hold the same probe image, with the size of $b<a(M-a)$,
while the second term trains on a quadruplet set that contains two pairs with different probe images.

\noindent \textbf{Triplet vs. Contrastive:}
Compared the triplet loss in Eq.~\ref{eq:learnedtripletloss} and the first term of the contrastive loss in Eq.~\ref{eq:convertsets},
It can be seen that the only difference is the threshold $u$ in $max[u,g(x_i,x_j)^2\!+\!\alpha_{cts}\!-\!g(x_l,x_k)^2]$.
The triplet loss purely considers the error (\ie $g(x_i,x_j)^2\!+\!\alpha_{cts}\!-\!g(x_l,x_k)^2$) of the relative distance between positive and negative pairs as long as its exists ($>0$).
But the first term of the contrastive loss gives priority to the absolute distance of positive pairs $g(x_i,x_j)$ when the error of the relative distance is not large enough. It would cause the contrastive loss to obtain a small positive distance with the risk of existing errors in the relative distances between positive and negative pairs.

\noindent \textbf{Quadruplet vs. Contrastive:}
Then we compare our quadruplet loss in Eq.~\ref{eq:quadrupletloss} with the contrastive loss.
Besides the difference of the threshold $u$ in $max(u,\cdot)$,
we can find that the two margin thresholds in Eq.~\ref{eq:quadrupletloss} are different.
In the contrastive loss of Eq.~\ref{eq:convertsets}, the two terms share the same margin threshold $\alpha_{cts}$,
which indicates that the second term plays an equally important role as the first term.
It causes the contrastive loss prefers to the model with a low misclassification rate,
no matter whether the misclassifications come from orders with the same probe images or with different probe images.
This problem is ubiquitous in binary classifiers.
But in person ReID, what we care most is those with the same probe.
This setting would lead the trained model to an undesired solution.

An example is shown in Fig.~\ref{fig:distribution2}.
Case 1 and 2 illustrate two projected distributions of scores obtained by binary classifiers
containing images from three persons (person A, B and C).
For each pair sample, the score underneath is a probability denoting the similarity between its two images.
Probe:\textit{X} indicates where an image from person \textit{X} is used as a probe image (the left image in a pair).
For example, Probe:A means an image from person A is used as a probe image.
The green-coloured rectangle indicates a positive pair, and the red rectangle for the negative pair.
In Case 1, it is evident that for each probe image (w.r.t one particular person),
we can get the correct rank-1 result.
However, in this case it is very difficult for a classifier to determine a suitable threshold to distinguish positive and negative pairs
(\eg, less than two misclassified samples).
On the contrary in Case 2, where the vertical dashed line denotes the decision threshold learned by the classifier,
the classifier has a lower misclassification rate.
As a result, a binary classifier in Eq.~\ref{eq:convertsets} will favor Case 2 rather than Case 1,
as the binary classification loss in Case 2 (one misclassified sample) will be lower than that in Case 1.
But in person ReID, we prefer Case 1, which outputs correct rank-1 results for all of the three persons,
rather than Case 2 that contains a false rank-1 result.

In our quadruplet loss, we treat two terms in Eq.~\ref{eq:quadrupletloss} differently to solve this problem\footnote{This can't be achieved in a traditional network with the binary classification loss, unless the input of the network changes from doublets into quadruplets as ours.},
which are trained with different margin thresholds.
The second term provides a relatively weaker auxiliary constraint,
while the first term maintains the stronger constraint and plays a dominant role.

\noindent \textbf{Quadruplet vs. Triplet:}
As shown in Section~\ref{sec:quadrupletloss}, the triplet loss is part of our quadruplet loss, but without the second term in Eq.~\ref{eq:quadrupletloss}. The second term provides a help from the perspective of orders with different probe images.
It can further enlarge the inter-class variations and improve the performance on the testing data.

As a result, we can find that our quadruplet loss covers the weaknesses from both the binary classification loss and the triplet loss to some extent,
and takes their advantages in person ReID which achieves a better performance than either of them.
In Section~\ref{ssec:experiment1}, we also provide related experiments to compare our quadruplet network with the traditional networks using the contrastive loss in Eq.~\ref{eq:learnedcontrastloss}.

\begin{figure}[!t]
\centering
\includegraphics[width=1\linewidth]{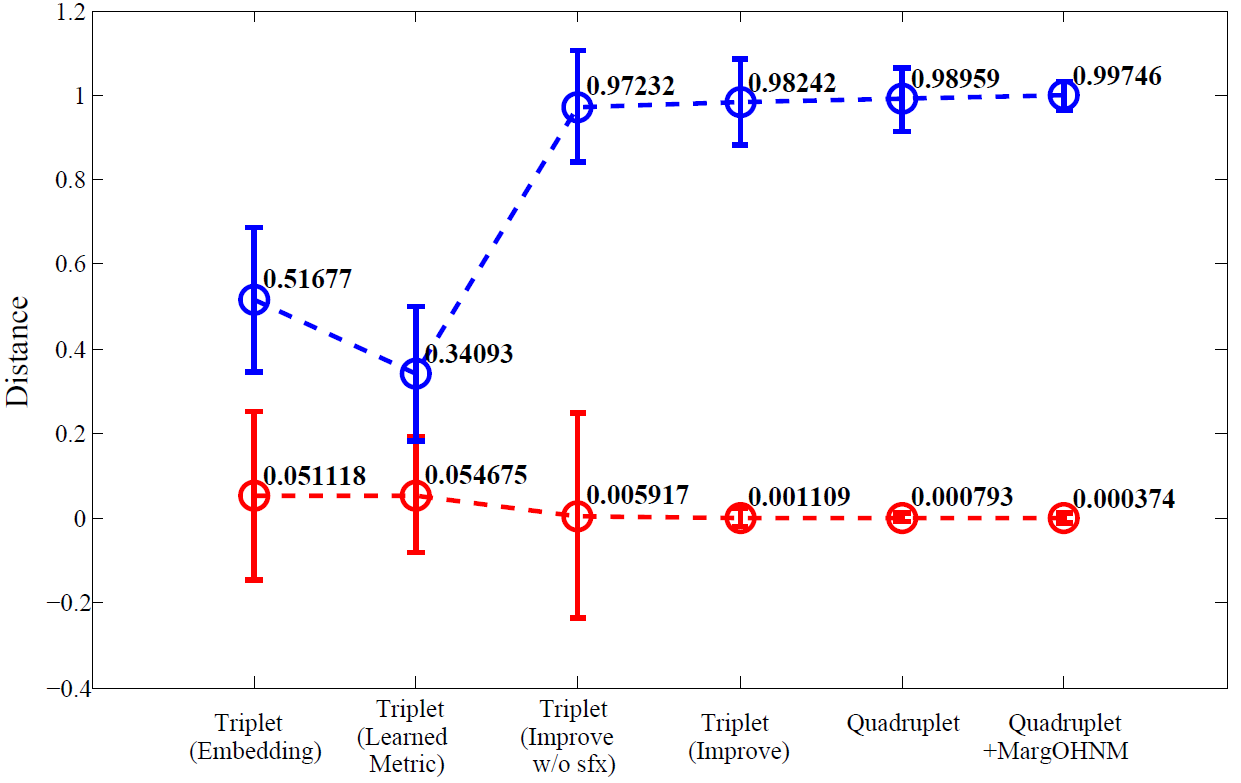}
\figupvspace
\caption{The distributions of intra- and inter- class distances from different models on CUHK03 training set. The red and blue lines indicate intra- and inter- distance respectively.}
\label{fig:intra_inter}
\figvspace
\end{figure}

\section{Experiment}
\label{sec:experiments}

We conduct two sets of experiments: 1) to evaluate the performance of different losses; 2) to compare the proposed approach with state-of-the-art methods.

\subsection{Implementation and Datasets}

Our method is implemented on the Caffe framework \cite{jia2014caffe}.
All images are resized to $227\times227$ before being fed to network.
The learning rate is set to $10^{-3}$, and the batch size is 128.
For all the datasets, we horizontally mirror each image
and increase the dataset sizes fourfold.
%During the margin-based hard negative mining, for the softmax layer, we select the top-32 samples to feedback.
As the marge-based hard negative mining is switched off,
the margin thresholds $\alpha_1$ and $\alpha_2$ in Eq.~\ref{eq:quadrupletloss} are set to 1 and 0.5 respectively.
In the beginning of the training on the margin-based hard negative mining,
the two distributions are chaos, and the average distance would be meaningless.
To provide an effective startup and accelerate the convergence,
we initialize the network with a pre-trained model on fixed margin thresholds.
For all other networks, we use a pre-trained AlexNet model (trained on Imagenet dataset~\cite{krizhevsky/nips2012imagenet})
to initialize the kernel weights of the first two convolutional layers.
Cumulative Matching Characteristics (CMC) curves are employed to measure the ReID performance.
We report the single-shot results on all the datasets.

The experiment is conducted on three datasets including CUHK03 \cite{Li/cvpr2014deepreid}, CUHK01~\cite{li/accv2012human},
and VIPeR~\cite{gray2007evaluating}.
The CUHK03 \cite{Li/cvpr2014deepreid} contains 13164 images from 1360 persons. We randomly select 1160 persons for training, 100 persons for validation and 100 persons for testing,
following exactly the same setting as \cite{Li/cvpr2014deepreid} and \cite{ahmed/cvpr2015improved}.
The CUHK01~\cite{li/accv2012human} and VIPeR~\cite{gray2007evaluating} datasets have 971 and 632 persons respectively, captured from two camera views. Every individual contains two images from each camera.
For VIPeR and CUHK01 dataset, the individuals are randomly divided into two equal parts, with one used for training and the other for testing.
%In CUHK01 dataset, we randomly choose 100 persons for testing, and the rest 871 persons are used for training.
%These settings consist with other methods~\cite{ahmed/cvpr2015improved,sircir2016cvpr,chen2015deep}.
%As there are also some methods dividing CUHK01 into two equal parts in their experiments, we also list them for a comprehensive view and mark their results with $*$ in Table~\ref{table:exp1}.
Note that for the comparison purpose, we further report our results on CUHK01 with another setting:
100 persons are randomly chosen for testing, and the rest 871 persons are used for training, denoted by \emph{CUHK01(p=100)}.

\begin{table*}
\renewcommand{\arraystretch}{0.85}
\caption{The CMC performance of the state-of-the-art methods and different architectures in our method on three representative datasets.}
\tableupvspace
\begin{center}
\resizebox{\linewidth}{!}{
\begin{tabular} {c||c|c|c||c|c|c||c|c|c||c|c|c}
  \hline
  \multirow{2}{*}{Method} & \multicolumn{3}{|c||}{CUHK03} & \multicolumn{3}{|c||}{CUHK01(p=486)} & \multicolumn{3}{|c||}{CUHK01(p=100)} & \multicolumn{3}{|c}{VIPeR} \\
  \cline{2-13}
   & $r=1$ & $r=5$ & $r=10$ & $r=1$ & $r=5$ & $r=10$ & $r=1$ & $r=5$ & $r=10$ & $r=1$ & $r=5$ & $r=10$ \\
  \hline
  %PRDC~\cite{zheng/cvpr2011person} & - & - & - & - & - & - & - & - & - & 15.70 & 38.40 & 53.90 \\
  %SDALF~\cite{farenzena/cvpr2010person} & 5.60 & 23.45 & 36.09 & 9.90 & 22.57 & 30.33 & 9.90 & 41.21 & 56.00 & 19.87 & 38.89 & 49.37 \\
  ITML~\cite{davis/icml2007information} & 5.53 & 18.89 & 29.96 & 15.98 & 35.22 & 45.60 & 17.10 & 42.31 & 55.07 & - & - & - \\
  %LMNN~\cite{weinberger2009distance} & 7.29 & 21.00 & 32.06 & 13.45 & 31.33 & 42.25 & 21.17 & 49.67 & 62.47 & - & - & - \\%
  eSDC~\cite{zhao/cvpr2013unsupervised} & 8.76 & 24.07 & 38.28 & 19.76 & 32.72 & 40.29 & 22.84 & 43.89 & 57.67 & 26.31 & 46.61 & 58.86 \\
  KISSME~\cite{Koestinger/cvpr2012scale} & 14.17 & 48.54 & 52.57 & - & - & - & 29.40 & 57.67 & 62.43 & 19.60 & 48.00 & 62.20 \\
  FPNN~\cite{Li/cvpr2014deepreid} & 20.65 & 51.00 & 67.00 & - & - & - & 27.87 & 64.00 & 77.00 & - & - & - \\
  mFilter~\cite{Zhao/cvpr2014midlevel} & - & - & - & 34.30 & 55.00 & 65.30 & - & - & - & 29.11 & 52.34 & 65.95 \\
  kLFDA~\cite{xiong/eccv2014person} & 48.20 & 59.34 & 66.38 & 32.76 & 59.01 & 69.63 & 42.76 & 69.01 & 79.63 & 32.33 & 65.78 & 79.72 \\
  DML~\cite{yi/icpr2014deep} & - & - & - & - & - & - & - & - & - & 34.40 & 62.15 & 75.89 \\
  IDLA~\cite{ahmed/cvpr2015improved} & 54.74 & 86.50 & 94.00 & 47.53 & 71.50 & 80.00 & 65.00 & 89.50 & 93.00 & 34.81 & 63.32 & 74.79 \\
  SIRCIR~\cite{sircir2016cvpr} & 52.17 & 85.00 & 92.00 & - & - & - & 72.50 & 91.00 & 95.50 & 35.76 & 67.00 & 82.50 \\
  DeepRanking~\cite{chen2015deep} & - & - & - & 50.41 & 75.93 & 84.07 & 70.94 & 92.30 & 96.90 & 38.37 & 69.22 & 81.33 \\
  DeepRDC~\cite{ding2015deep} & - & - & - & - & - & - & - & - & - & 40.50 & 60.80 & 70.40 \\
  NullReid~\cite{null2016cvpr} & 58.90 & 85.60 & 92.45 & 64.98 & 84.96 & 89.92 & - & - & - & 42.28 & 71.46 & 82.94 \\
  %SiameseLSTM~\cite{lstm2016cvpr} & 57.30 & 80.10 & 88.30 & - & - & - & - & - & - & 42.40 & 68.70 & 79.40 \\
  Ensembles~\cite{paisitkriangkrai/cvpr2015learning} & 62.10 & 89.10 & 94.30 & 53.40 & 76.30 & 84.40 & - & - & - & 45.90 & 77.50 & \textbf{88.90} \\
  DeepLDA~\cite{LDA2016arXiv} & 63.23 & 89.95 & 92.73 & - & - & - & 67.12 & 89.45 & 91.68 & 44.11 & 72.59 & 81.66 \\
  GOG~\cite{hierarchical2016cvpr} & 67.30 & 91.00 & 96.00 & 57.80 & 79.10 & 86.20 & - & - & - & \textbf{49.70} & \textbf{79.70} & 88.70 \\
  GatedSiamese~\cite{gated2016cvpr} & 68.10 & 88.10 & 94.60 & - & - & - & - & - & - & 37.80 & 66.90 & 77.40 \\
  ImpTrpLoss~\cite{imptrp2016cvpr} & - & - & - & 53.70 & 84.30 & 91.00 & - & - & - & 47.80 & 74.70 & 84.80 \\
  DGD~\cite{dgd2016cvpr} & \textbf{80.50} & 94.90 & 97.10 & \textbf{71.70} & \textbf{88.60} & \textbf{92.60} & - & - & - & 35.40 & 62.30 & 69.30 \\
  \hline
  \textbf{BL1:} Triplet(Embedding) & 60.13 & 90.51 & 95.15 & 44.24 & 67.08 & 77.57 & 63.50 & 80.00 & 89.50 & 28.16 & 52.22 & 65.19 \\
  \textbf{BL2:} Triplet(Learned Metric) & 61.60 & 92.41 & 97.47 & 58.74 & 80.35 & 88.07 & 77.00 & 94.00 & 97.50 & 40.19 & 70.25 & 82.91 \\
  Triplet(Improved w/o sfx) & 70.25 & 95.97 & 98.10 & 58.85 & 82.61 & 88.37 & 77.50 & 95.00 & 96.50 & 44.30 & 72.47 & 80.06 \\
  Triplet(Improved) & 72.78 & 95.97 & 97.68 & 59.26 & 82.41 & 88.27 & 78.00 & 95.50 & 98.00 & 44.30 & 71.84 & 81.96 \\
  Quadruplet & 74.47 & \textbf{96.62} & 98.95 & 62.55 & 83.02 & 88.79 & 79.00 & 96.00 & 97.00 & 48.42 & 74.05 & 84.49 \\
  \textbf{BL3:} Classification & 68.35 & 93.46 & 97.47 & 58.74 & 79.01 & 87.14 & 76.50 & 94.00 & 97.00 & 44.30 & 69.94 & 81.96 \\
  Quadruplet + MargOHNM & 75.53 & 95.15 & \textbf{99.16} & 62.55 & 83.44 & 89.71 & \textbf{81.00} & \textbf{96.50} & \textbf{98.00} & 49.05 & 73.10 & 81.96 \\
  \hline
\end{tabular}}
\end{center}
\label{table:exp1}
\tablevspace
\end{table*}

\subsection{Results of Quadruplet Network}
\label{ssec:experiment1}

\noindent \textbf{Different Losses.}
We conduct experiments with different losses and provide several baselines
to illustrate the effectiveness of each component in our method.
Results are shown in Table~\ref{table:exp1}.
There are three baselines.
The first two baselines are the networks in Fig.~\ref{fig:nets} (a) and (b)
using a triplet loss with an embedding Euclidean distance and a learned metric respectively,
denoted by \textit{BL1:Triplet(Embedding)} and \textit{BL2:Triplet(Learned Metric)}.
The third one is a traditional network using a binary classification loss mentioned in Section.~\ref{sec:relationship}
with the same eight layers as our framework, denoted by \textit{BL3:Classification}.
Our improved triplet loss containing a normalization with a two-dimensional output in Fig.~\ref{fig:nets} (c) is denoted by \textit{Triplet(Improved)},
and \textit{Triplet(Improved w/o sfx)} means that without the help of the softmax loss.
The network \textit{Quadruplet} indicates the proposed quadruplet network in Fig.~\ref{fig:framework}.
Compared our \textit{Triplet(Improved)} with two baselines (\textit{BL1:Triplet(Embedding)} and \textit{BL2:Triplet(Learned Metric)}),
it's obvious that the learned similarity metric with a two-dimensional output is better than
the embedding one or that with a one-dimensional output like Wang's~\cite{sircir2016cvpr}.
When comparing the performance between \textit{Triplet(Improved w/o sfx)} and \textit{Triplet(Improved)}, adding the softmax loss could slightly boost the overall performance of our improved triplet loss.
And if the new constraint is brought in, for all three datasets,
the performance of \textit{Quadruplet} is consistently better than \textit{Triplet(Improved)},
which implies the effectiveness of our proposed quadruplet loss.
What's more, as said in Section~\ref{sec:relationship},
our quadruplet loss has connections with the binary classification loss.
From the comparison between \textit{Quadruplet} and the baseline \textit{BL3:Classification},
it can be found that our quadruplet loss can overcome the weakness of the binary classification loss and produce a great improvement on the performance.

\noindent \textbf{With vs without margin-based hard negative mining}.
Then we test the effectiveness of our margin-based hard negative mining.
In Table~\ref{table:exp1}, the term \textit{+MargOHNM} indicates the network using our margin-based online hard negative mining.
It is obvious that when the \textit{+MargOHNM} is used,
the results of \textit{Quadruplet+MargOHNM} are further improved,
which suggest that the margin-based online hard negative mining can select samples effectively and enhance the performance.
%For the VIPeR and CUHK01 dataset, the improvement of \textit{MargOHNM} is not as obvious as that in CUHK03.
%It shows that the effectiveness of our margin-based hard negative mining increases as the training set becomes larger and more difficult.
It can be seen that \textit{+MargOHNM} performs better for rank-n (n$>$1) in CUHK03 and CUHK01, but on the opposite way in VIPeR.
As we adopt the mean values of two learned distributions to replace the margins.
The confidences of two distributions have a great influence on the results of \textit{+MargOHNM}.
For CUHK03 and CUHK01, the performance (\ie learned distributions) is on a high-confidence level (rank1~70\%+), much higher than that of VIPeR (rank1~40\%+). As a result, \textit{+MargOHNM} can work better on CUHK03 and CUHK01.

\noindent \textbf{Effects on intra- and inter- class variations.}
We also provide the distributions of intra- and inter- class distances from models trained with different losses on CUHK03 training set in Fig.~\ref{fig:intra_inter}.
As the distances from \textit{BL2:Triplet(Learned Metric)} do not range from 0 to 1, we normalize the distances into [0,1] and get the results.
From Fig.~\ref{fig:intra_inter}, we can see that our \textit{Triplet (Improved)}, \textit{Quadruplet} and \textit{Quadruplet+MargOHNM} gradually make the average intra-class distance smaller and smaller, and make the average inter-class distance larger and larger. For the large intra-class distance and the small inter-class distance of \textit{BL2:Triplet(Learned Metric)},
that's due to the lack of normalization on the output layers as said in Section~\ref{ssec:tripletloss}.

\subsection{Comparison with the state of the arts}
\label{ssec:experiment2}

We compare ours with representative ReID methods including 18 algorithms.
In Table \ref{table:exp1}, it is noted that our results are better than most approaches above,
which further confirm the effectiveness of our proposed method.
Under the rank-1 accuracy, our multi-task network outperforms most of existing person ReID algorithms on all three datasets.
The DGD~\cite{dgd2016cvpr} achieves better performance than us, but
it combines all current datasets together as its training data which is much larger than ours.
Even so, our rank-n (n$>$1) performance on CUHK03 is higher than DGD's.
The loss in DGD is designed for maximizing the top-1 classification accuracy, with less emphasis on top-n (n$>$1) accuracies.
The top-1 classification accuracy corresponds to the rank-1 result.
Our quadruplet loss cares both the ranking orders and the rank-1 accuracy,
that's why our method outperforms DGD in rank-n (n$>$1) though not better in terms of rank-1.
Since VIPeR is relatively small, it is expected that deep learning might not be demonstrated to reach its full potential; instead, a hand-crafted metric learning may be more advantageous on this set, like GOG~\cite{hierarchical2016cvpr} and Ensembles~\cite{paisitkriangkrai/cvpr2015learning}.
It is noted that DeepLDA~\cite{LDA2016arXiv} and ImpTrpLoss~\cite{imptrp2016cvpr} also focus on the intra- and inter- class variations like us, as mentioned in Section~\ref{sec:relatedwork}.
From the results compared with DeepLDA~\cite{LDA2016arXiv} and ImpTrpLoss~\cite{imptrp2016cvpr},
we can conclude that our constraint is more effective than theirs.

\section{Conclusion}
\label{sec:conclusion}

In this paper, a quadruplet loss is proposed to handle the weakness of the triplet loss on person ReID.
And a quadruplet network using a margin-based online hard negative mining is presented based on the quadruplet loss,
which has outperformed most of the state-of-the-art methods on CUHK03, CUHK01 and VIPeR.

%\section*{Acknowledgement}
%\label{sec:acknowledgement}
%This work is funded by the National Key Research and Development Program of China (2016YFB1001005), the National Natural Science Foundation of China (Grant No. 61673375 and Grant No. 61403383), and the Projects of Chinese Academy of Science, (Grant No. QYZDB-SSW-JSC006, Grant No.173211KYSB20160008).

{\small
\bibliographystyle{ieee}
\bibliography{egbib}

\begin{thebibliography}{10}\itemsep=-1pt

\bibitem{ahmed/cvpr2015improved}
E.~Ahmed, M.~Jones, and T.~K. Marks.
\newblock An improved deep learning architecture for person re-identification.
\newblock In {\em {CVPR}}, 2015.

\bibitem{chen2015deep}
S.-Z. Chen, C.-C. Guo, and J.-H. Lai.
\newblock Deep ranking for person re-identification via joint representation
  learning.
\newblock {\em {TIP}}, 25(5):2353--2367, 2016.

\bibitem{chen2017aaai}
W.~Chen, X.~Chen, J.~Zhang, and K.~Huang.
\newblock A multi-task deep network for person re-identification.
\newblock In {\em {AAAI}}, 2017.

\bibitem{imptrp2016cvpr}
D.~Cheng, Y.~Gong, S.~Zhou, J.~Wang, and N.~Zheng.
\newblock Person re-identification by multi-channel parts-based cnn with
  improved triplet loss function.
\newblock In {\em {CVPR}}, 2016.

\bibitem{cortes1995support}
C.~Cortes and V.~Vapnik.
\newblock Support-vector networks.
\newblock {\em Machine learning}, 20(3):273--297, 1995.

\bibitem{davis/icml2007information}
J.~V. Davis, B.~Kulis, P.~Jain, S.~Sra, and I.~S. Dhillon.
\newblock Information-theoretic metric learning.
\newblock In {\em {ICML}}, 2007.

\bibitem{ding2015deep}
S.~Ding, L.~Lin, G.~Wang, and H.~Chao.
\newblock Deep feature learning with relative distance comparison for person
  re-identification.
\newblock {\em Pattern Recognition}, 48(10):2993--3003, 2015.

\bibitem{gou2016person}
M.~Gou, X.~Zhang, A.~Rates-Borras, S.~Asghari-Esfeden, M.~Sznaier, and
  O.~Camps.
\newblock Person re-identification in appearance impaired scenarios.
\newblock In {\em {BMVC}}, 2016.

\bibitem{gray2007evaluating}
D.~Gray, S.~Brennan, and H.~Tao.
\newblock Evaluating appearance models for recognition, reacquisition, and
  tracking.
\newblock In {\em Proc. IEEE International Workshop on Performance Evaluation
  for Tracking and Surveillance (PETS)}, 2007.

\bibitem{ctsloss2006cvpr}
R.~Hadsell, S.~Chopra, and Y.~LeCun.
\newblock Dimensionality reduction by learning an invariant mapping.
\newblock In {\em {CVPR}}, 2006.

\bibitem{hirzer2012avss}
M.~Hirzer, P.~M. Roth, and H.~Bischof.
\newblock Person re-identification by efficient impostor-based metric learning.
\newblock In {\em Advanced Video and Signal-Based Surveillance (AVSS), 2012
  IEEE Ninth International Conference on}, 2012.

\bibitem{jia2014caffe}
Y.~Jia, E.~Shelhamer, J.~Donahue, S.~Karayev, J.~Long, R.~Girshick,
  S.~Guadarrama, and T.~Darrell.
\newblock Caffe: Convolutional architecture for fast feature embedding.
\newblock In {\em ACM on Multimedia}, 2014.

\bibitem{karanam2016comprehensive}
S.~Karanam, M.~Gou, Z.~Wu, A.~Rates-Borras, O.~Camps, and R.~J. Radke.
\newblock A comprehensive evaluation and benchmark for person
  re-identification: Features, metrics, and datasets.
\newblock {\em arXiv preprint arXiv:1605.09653}, 2016.

\bibitem{Koestinger/cvpr2012scale}
M.~Koestinger, M.~Hirzer, P.~Wohlhart, P.~M. Roth, and H.~Bischof.
\newblock Large scale metric learning from equivalence constraints.
\newblock In {\em {CVPR}}, 2012.

\bibitem{krizhevsky/nips2012imagenet}
A.~Krizhevsky, I.~Sutskever, and G.~E. Hinton.
\newblock Imagenet classification with deep convolutional neural networks.
\newblock In {\em {NIPS}}, 2012.

\bibitem{dangwei2017cvpr}
D.~Li, X.~Chen, Z.~Zhang, and K.~Huang.
\newblock Learning deep context-aware features over body and latent parts for
  person re-identification.
\newblock In {\em {CVPR}}, 2017.

\bibitem{Li/cvpr2013locally}
W.~Li and X.~Wang.
\newblock Locally aligned feature transforms across views.
\newblock In {\em {CVPR}}, 2013.

\bibitem{li/accv2012human}
W.~Li, R.~Zhao, and X.~Wang.
\newblock Human reidentification with transferred metric learning.
\newblock In {\em {ACCV}}, 2012.

\bibitem{Li/cvpr2014deepreid}
W.~Li, R.~Zhao, T.~Xiao, and X.~Wang.
\newblock Deepreid: Deep filter pairing neural network for person
  re-identification.
\newblock In {\em {CVPR}}, 2014.

\bibitem{li/cvpr2013learning}
Z.~Li, S.~Chang, F.~Liang, T.~S. Huang, L.~Cao, and J.~R. Smith.
\newblock Learning locally-adaptive decision functions for person verification.
\newblock In {\em {CVPR}}, 2013.

\bibitem{liao2015person}
S.~Liao, Y.~Hu, X.~Zhu, and S.~Z. Li.
\newblock Person re-identification by local maximal occurrence representation
  and metric learning.
\newblock In {\em {CVPR}}, 2015.

\bibitem{Liao/iccv2015psd}
S.~Liao and S.~Z. Li.
\newblock Efficient psd constrained asymmetric metric learning for person
  re-identification.
\newblock In {\em {ICCV}}, 2015.

\bibitem{hierarchical2016cvpr}
T.~Matsukawa, T.~Okabe, E.~Suzuki, and Y.~Sato.
\newblock Hierarchical gaussian descriptor for person re-identification.
\newblock In {\em {CVPR}}, 2016.

\bibitem{paisitkriangkrai/cvpr2015learning}
S.~Paisitkriangkrai, C.~Shen, and A.~van~den Hengel.
\newblock Learning to rank in person re-identification with metric ensembles.
\newblock In {\em {CVPR}}, 2015.

\bibitem{pedagadi/cvpr2013local}
S.~Pedagadi, J.~Orwell, S.~Velastin, and B.~Boghossian.
\newblock Local fisher discriminant analysis for pedestrian re-identification.
\newblock In {\em {CVPR}}, 2013.

\bibitem{schroff2015facenet}
F.~Schroff, D.~Kalenichenko, and J.~Philbin.
\newblock Facenet: A unified embedding for face recognition and clustering.
\newblock In {\em {CVPR}}, 2015.

\bibitem{Shen/iccv2015structure}
Y.~Shen, W.~Lin, J.~Yan, M.~Xu, J.~Wu, and J.~Wang.
\newblock Person re-identification with correspondence structure learning.
\newblock In {\em {ICCV}}, 2015.

\bibitem{su2015multi}
C.~Su, F.~Yang, S.~Zhang, Q.~Tian, L.~S. Davis, and W.~Gao.
\newblock Multi-task learning with low rank attribute embedding for person
  re-identification.
\newblock In {\em {ICCV}}, 2015.

\bibitem{deepattr2016arXiv}
C.~Su, S.~Zhang, J.~Xing, W.~Gao, and Q.~Tian.
\newblock Deep attributes driven multi-camera person re-identification.
\newblock In {\em {ECCV}}, 2016.

\bibitem{gated2016cvpr}
R.~R. Varior, M.~Haloi, and G.~Wang.
\newblock Gated siamese convolutional neural network architecture for human
  re-identification.
\newblock In {\em {ECCV}}, 2016.

\bibitem{lstm2016cvpr}
R.~R. Varior, B.~Shuai, J.~Lu, D.~Xu, and G.~Wang.
\newblock A siamese long short-term memory architecture for human
  re-identification.
\newblock In {\em {ECCV}}, 2016.

\bibitem{sircir2016cvpr}
F.~Wang, W.~Zuo, L.~Lin, D.~Zhang, and L.~Zhang.
\newblock Joint learning of single-image and cross-image representations for
  person re-identification.
\newblock In {\em {CVPR}}, 2016.

\bibitem{Wang2014cvpr}
J.~Wang, Y.~Song, T.~Leung, C.~Rosenberg, J.~Wang, J.~Philbin, B.~Chen, and
  Y.~Wu.
\newblock Learning fine-grained image similarity with deep ranking.
\newblock In {\em {CVPR}}, 2014.

\bibitem{wen2016discriminative}
Y.~Wen, K.~Zhang, Z.~Li, and Y.~Qiao.
\newblock A discriminative feature learning approach for deep face recognition.
\newblock In {\em {ECCV}}, 2016.

\bibitem{personnet2016arXiv}
L.~Wu, C.~Shen, and A.~v.~d. Hengel.
\newblock Personnet: person re-identification with deep convolutional neural
  networks.
\newblock {\em arXiv preprint arXiv:1601.07255}, 2016.

\bibitem{LDA2016arXiv}
L.~Wu, C.~Shen, and A.~van~den Hengel.
\newblock Deep linear discriminant analysis on fisher networks: A hybrid
  architecture for person re-identification.
\newblock {\em Pattern Recognition}, 2016.

\bibitem{dgd2016cvpr}
T.~Xiao, H.~Li, W.~Ouyang, and X.~Wang.
\newblock Learning deep feature representations with domain guided dropout for
  person re-identification.
\newblock In {\em {CVPR}}, 2016.

\bibitem{xiong/eccv2014person}
F.~Xiong, M.~Gou, O.~Camps, and M.~Sznaier.
\newblock Person re-identification using kernel-based metric learning methods.
\newblock In {\em {ECCV}}, 2014.

\bibitem{yang2016metric}
Y.~Yang, Z.~Lei, S.~Zhang, H.~Shi, and S.~Z. Li.
\newblock Metric embedded discriminative vocabulary learning for high-level
  person representation.
\newblock In {\em {AAAI}}, 2016.

\bibitem{yang2017aaai}
Y.~Yang, S.~Z. Li, L.~Wen, and S.~Lyu.
\newblock Unsupervised learning of multi-level descriptors for person
  re-identification.
\newblock In {\em {AAAI}}, 2017.

\bibitem{yi/icpr2014deep}
D.~Yi, Z.~Lei, S.~Liao, and S.~Z. Li.
\newblock Deep metric learning for person re-identification.
\newblock In {\em {ICPR}}, 2014.

\bibitem{null2016cvpr}
L.~Zhang, T.~Xiang, and S.~Gong.
\newblock Learning a discriminative null space for person re-identification.
\newblock In {\em {CVPR}}, 2016.

\bibitem{zhao/cvpr2013unsupervised}
R.~Zhao, W.~Ouyang, and X.~Wang.
\newblock Unsupervised salience learning for person re-identification.
\newblock In {\em {CVPR}}, 2013.

\bibitem{Zhao/cvpr2014midlevel}
R.~Zhao, W.~Ouyang, and X.~Wang.
\newblock Learning mid-level filters for person re-identification.
\newblock In {\em {CVPR}}, 2014.

\bibitem{zheng2016person}
L.~Zheng, Y.~Yang, and A.~G. Hauptmann.
\newblock Person re-identification: Past, present and future.
\newblock {\em arXiv preprint arXiv:1610.02984}, 2016.

\end{thebibliography}
}

\end{document}